# Enabling Shared-Control for A Riding Ballbot System

Yu Chen, Mahshid Mansouri, Chenzhang Xiao, student member, IEEE, Ze Wang, Elizabeth T. Hsiao-Wecksler, member, IEEE, William R. Norris, member, IEEE

*Abstract*— **This study introduces a shared-control approach for collision avoidance in a self-balancing riding ballbot, called PURE, marked by its dynamic stability, omnidirectional movement, and hands-free interface. Integrated with a sensor array and a novel Passive Artificial Potential Field (PAPF) method, PURE provides intuitive navigation with deceleration assistance and haptic/audio feedback, effectively mitigating collision risks. This approach addresses the limitations of traditional APF methods, such as control oscillations and unnecessary speed reduction in challenging scenarios. A human-robot interaction experiment, with 20 manual wheelchair users and able-bodied individuals, was conducted to evaluate the performance of indoor navigation and obstacle avoidance with the proposed shared-control algorithm. Results indicated that shared-control significantly reduced collisions and cognitive load without affecting travel speed, offering intuitive and safe operation. These findings highlight the shared-control system's suitability for enhancing collision avoidance in self-balancing mobility devices, a relatively unexplored area in assistive mobility research.**

## I. INTRODUCTION

Navigating crowded spaces with a mobility device demands significant mental effort. The development of "smart" powered wheelchairs [1], [2] highlights the need to assist individuals with physical or cognitive challenges in safely operating these devices. While early efforts focused on fully autonomous wheelchairs [1], the complexity of navigating varied environments and the need to preserve user independence led to shared-control systems [3]. These systems balance human decision-making with robotic motion control and collision avoidance, reducing the user's cognitive and physical burden, and allowing them to focus on reaching their destination or engaging in social activities rather than on navigation [4].

Shared-control systems can be classified into deliberative methods, like Rapidly Exploring Random Trees (RRT) [5], Model Predictive Control (MPC) [6], Reinforcement Learning (RL) and Bayesian methods [7], and reactive methods, such as Vector Field Histograms (VFH) [8], Dynamic Window Approach (DWA) [9], and Artificial Potential Fields (APF) [4], [10], [11]. While each has its merits, this study focused on a reactive shared-control paradigm to promote user independence and leverage human abstract reasoning.

Our research group has developed PURE (Personalized Unique Rolling Experience), a novel mobility device featuring a ballbot drivetrain — a self-balancing mobile robot that rides atop a ball [12], [13], [14]. PURE's instrumented seat and wearable sensor enable intuitive, lean-to-steer hands-free control, omnidirectional motion, dynamic stability, and a compact form [12], [15], [16]. These features make it ideal for navigating narrow indoor environments designed for human access. The device's omnidirectional maneuverability and under-actuated dynamics necessitate heightened awareness for collision avoidance, posing challenges for estimating braking distances, especially for novice users in congested areas. By adopting a shared-control paradigm, this study aims to enhance the safety of the PURE system and broaden the inclusiveness and accessibility of this mobility technology to a diverse range of users.

Developing a shared-control paradigm for self-balancing systems is non-trivial. Conventional braking systems cannot be used on a dynamically stable platform because halting the actuators would disrupt the device's ability to maintain balance through dithering. Instead, safe deceleration must be achieved by tilting the device in the opposite direction. Moreover, unlike traditional joystick-controlled wheelchairs, where the rider's command signal can be completely overridden, the passive compliance of self-balancing systems allows the rider to force movement by significantly shifting their center of mass.

To address these challenges, this study proposed a Passive Artificial Potential Field (PAPF) method, building on traditional APF formulations [17]. While APF-based shared-control methods have been applied successfully in contexts like intelligent walkers [18] and teleoperations of wheeled humanoids [19], their application to self-balancing mobility devices remains unexplored. This study demonstrated that in self-balancing devices like PURE, motion correction and haptic feedback are inherently linked, necessitating a revised APF approach to manage the device's unique dynamics and interactions.

The PAPF approach was designed to minimize collision risks through deceleration assistance and haptic feedback, tailored to the dynamics of self-balancing mobility devices. To evaluate this approach, a human-robot interaction (HRI) study was conducted with 20 participants, including manual wheelchair users (mWCUs) and able-bodied users (ABUs). Participants navigated PURE through complex obstacle courses with and without shared control, assessing improvements in performance, reductions in operational effort, and differences in user experience.

Key contributions of this research include a) the integration

Research supported by NSF NRI-2.0 #2024905.

Yu Chen*, Mahshid Mansouri, Chenzhang Xiao, Ze Wang, and Elizabeth T. Hsiao-Wecksler are with the Department of Mechanical Science & Engineering, University of Illinois Urbana-Champaign, Urbana, IL 61801 USA (e-mail: yuc6/mm64/cxiao3/zew2/ethw@illinois.edu)

William R. Norris is with the Department of Industrial & Enterprise Systems Engineering, University of Illinois Urbana-Champaign, Urbana, IL 61801 USA (e-mail: wrnorris@illinois.edu). *Corresponding author of a vision-guided perception system and a high-level controller for real-time shared control, b) the introduction of the PAPF method for enhanced safety in self-balancing devices, and c) comprehensive HRI evaluations to assess the proposed paradigm and gather feedback for future development. The evaluation aimed to answer the following research questions: 1) Does the performance improve when using the shared-controller? 2) Does the operational effort decrease when using the shared-controller? 3) Are there performance and preference differences between ABUs and mWCUs with the use of the shared-controller?

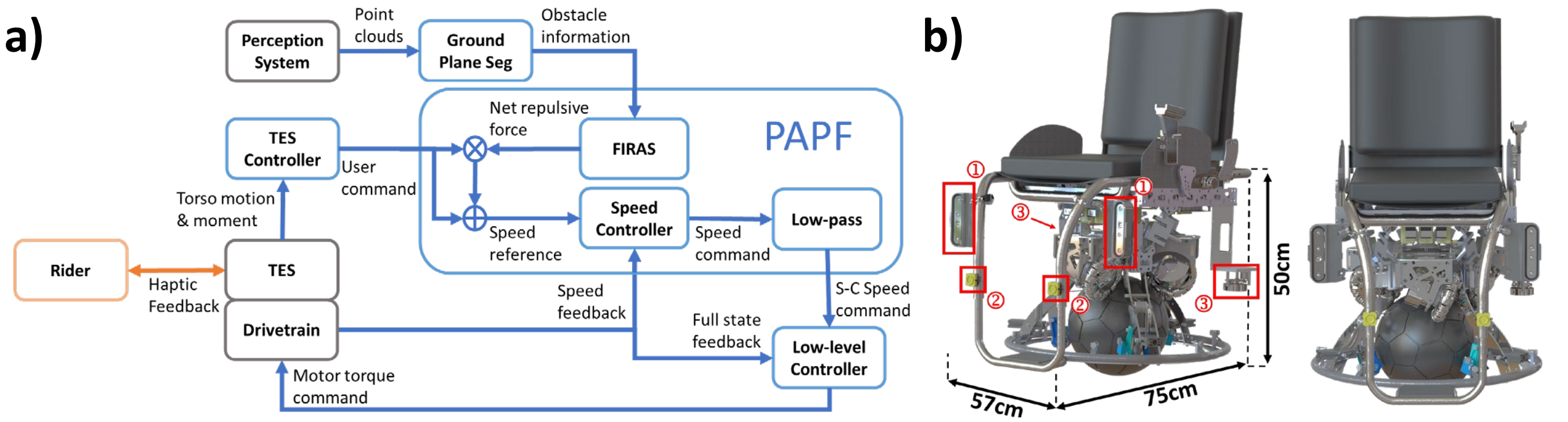

Fig.1. a) PURE control system architecture. b) PURE perception system sensor placement. ①RGB-D camera, ②ToF sensor, and ③LiDAR.

## II. Methodology

### A. System Specifications

The PURE prototype is designed for compactness and safety [12], with a footprint similar to a seated person and weighing approximately 37 kg. It offers a runtime of 2.5 to 4 hours and a top speed of 2.4 m/s. PURE's hardware comprises three main electronic modules (Fig.1a in grey color): 1) A ballbot drivetrain which manages system balance and velocity tracking [12], providing real-time state feedback for the shared-controller. 2) A Torso-dynamics Estimation System (TES) [15] which measures the rider's upper-body movements and later maps them into velocity control signals. 3) A robot perception system that ensures environmental awareness and collision prevention is hosted by a Single Board Computer (SBC, UP Xtreme i7) running Ubuntu 20.04 and ROS Noetic. PURE's perception system integrates two RGB-D cameras (RealSense D455), two Time-of-Flight (ToF) sensors (TeraRanger Evo Mini), and two single-beam LiDAR sensors (YDLIDAR X2) (Fig.1b) for a near 360° obstacle detection capability effectively up to 7 m. The SBC is connected to a speaker, providing an audio interface and issuing alarms when the chair approaches obstacles. Further details can be found in [20].

### B. Control Architecture

TES uses a Force Sensing Seat (FSS) with six uniaxial load cells for rider-seat interaction torque (translational) measurement and a chest-mounted IMU for upper body twist (yaw) detection [15]. The TES controller maps these signals to velocity commands. Although PURE's passive compliance makes it naturally drivable even without the help of the TES, augmentations of the rider's torso control signal through the IMU and FSS are necessary for riders who have limited torso motion function. Additionally, spin control is only attainable using the torso yaw signal.

PURE's perception system aggregates data from all sensors into point clouds. These are then processed by a ground plane segmentation node [21], which effectively extracts obstacle information from the surroundings. The shared-control system adjusts user commands based on sensory data and system velocity to minimize occurrence of undesired collisions.

The shared-control speed commands are fed into the low-level controller, which uses an LQR-PI cascaded control scheme [12] to manage PURE's movements. The dynamics in the frontal and sagittal planes are modeled using a wheeled-inverted pendulum approach.

### C. Passive Artificial Potential Field Shared-Control

In the conventional APF setup, the robot's motion is determined by the sum of the attractive force from the goal and the repulsive force from the obstacles [17]. However, in the context of shared-control, where the long-term motion goal is determined by the rider, only repulsive force is considered. This repulsive force can be calculated using the FIRAS function [17], which was defined as:

$$f^{APF} = \begin{cases} \eta\left(\frac{1}{\delta} - \frac{1}{\delta_{thre}}\right)^2 & \text{if } \delta \le \delta_{thre} \\ 0 & \text{if } \delta > \delta_{thre} \end{cases} \quad (1)$$

where $\delta_{thre}$ is a distance threshold representing the obstacle's radius of influence (m), $\delta$ is the minimum distance between the robot and the obstacle (m), and $\eta$ is a scaling factor that can be adjusted for each axis. The quadratic behavior of the function ensures that the device exerts maximum effort to avoid collision when near obstacles. Since it is challenging to accurately estimate the shape and size of obstacles and compute $\delta$, a simple adaptation is to treat all distance measurements as tiny obstacles. The overall repulsive force is the average of all the forces contributed by $N$ obstacle points calculated using the relative distance $\delta_n$ and the relative angle $\alpha_n$ with respect to the positive x-axis (front) of the robot:

$$f^{APF} = \begin{cases} \frac{1}{N}\sum_{n=1}^{N} g(\alpha_n)\,\eta\left(\frac{1}{\delta_n} - \frac{1}{\delta_{thre}}\right)^2 & \text{if } \delta \le \delta_{thre} \\ 0 & \text{if } \delta > \delta_{thre} \end{cases}$$

$$g(\alpha_n) = \begin{cases} \cos(\alpha_n) & \text{for } f_x^{APF} \\ \sin(\alpha_n) & \text{for } f_y^{APF} \end{cases} \quad (2)$$

Different from conventional APF, where an obstacle locates to the front-right (or front-left) of the robot generates repulsive forces in both longitudinal and lateral directions, PAPF calculates the repulsive forces $f_x^{APF}$ and $f_y^{APF}$ independently, considering the relative position of the obstacle. Thus, if an obstacle does not directly overlap with the robot's frontal area, the algorithm generates only lateral repulsive force, preventing unwanted slowdowns and facilitating smoother navigation in confined spaces. Additionally, PAPF considers only obstacle points within the path of motion, defined by the semicircle of the command velocity vector. It resists the rider's motion towards obstacles but does not introduce acceleration or proactively push the device away, thereby eliminating control oscillations and avoiding unnecessary movements when the ballbot is stationary or moving parallel to objects.

There are many ways to incorporate $f^{APF}$ into the robot's motion control loop. In this study, $f^{APF}$ is first saturated between $[-1, 1]$ and then scaled by the normalized norm of user commanded velocity signal, $v^{usr} \in [-1,1]$, to ensure it is only effective when the rider is driving. It is then added to the user commanded velocity to form the updated velocity command $v^{ideal}$. This approach of using $f^{APF}$ as a commanded velocity correction term is easier to implement and more robust against latency compared to injecting it into the torque control loop. Similar practices can be found in [19].

$$v^{ideal} = v^{usr} + (f^{APF}|v^{usr}|) \quad (3)$$

As mentioned earlier, the rider on PURE can continue to lean and drive toward an obstacle despite the shared-control's effort to slow down (Fig.2b). To address this issue, the shared-control system incorporates a cascaded speed tracking loop (Fig.2c). It compensates for undesired velocity by applying a counter lean and deceleration proportional to the unwanted velocity:

$$v^{comp} = v^{ideal} + \zeta(v^{ideal} - v^{fb}) \quad (4)$$

where $\zeta$ is a scaling factor that controlled the aggressiveness of the velocity tracking. It is set to zero if the feedback velocity $v^{fb}$ is higher than the targeted velocity to ensure that the shared-control does not accelerate the system. The counter lean that results from this tracking loop offset the COM shift caused by the rider and re-establishes a dynamically stable position. It also provides distinct haptic feedback to the rider and helpsconvey the shared-control's intention.

Finally, a low-pass filter is integrated into the system to mitigate the effects of noisy sensor measurements and rapid changes in the direction of motion. This low-pass filter smoothes out jitteriness by penalizing the difference between $v^{comp}$ and the commanded velocity from the last time step $v^{old}$, resulting in a more fluid and reliable movement.

$$v^{cmd} = v^{comp} + \epsilon(v^{old} - v^{comp}) \quad (5)$$

Note that, as PAPF calculates the forces for $x$ and $y$ axes (the front and left directions of the robot) separately, the above calculations are repeated for each axis.

In situations where PURE detects an imminent collision risk despite its efforts to slow down the rider, an audio alarm is activated to explicitly alert the rider to the danger. The volume of the alarm increases as the obstacle gets closer, providing a clear indication of the proximity of the collision.

The algorithm used in this study is available at: https://github.com/mszuyx/PURE_SMC

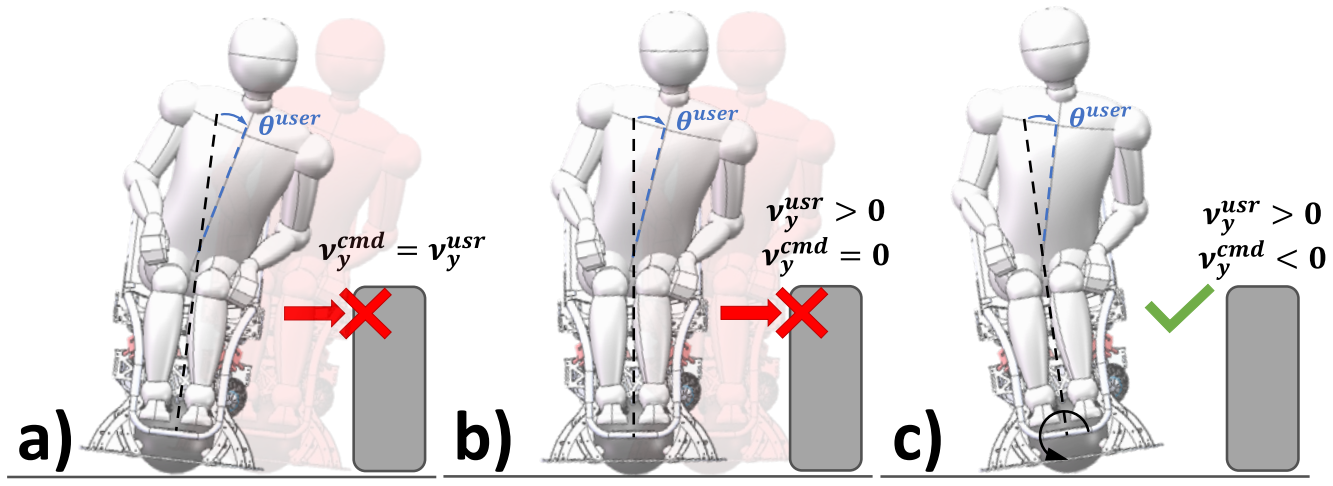

Fig.2. PURE's behaviors under same operator leaning motion and different control paradigms. a) No Shared-Control; b) Shared-Control without cascaded speed tracking; c) Shared-Control with cascaded speed tracking.

## III. Experimental Setup

### A. Subject Demographics

The study involved 20 participants, split evenly between manual wheelchair users (mWCUs) and able-bodied individuals (ABUs) (Table I). Participants ranged from 18 to 35 years old, weighed under 70 kg (prototype's payload limit), and had no visual impairments. mWCUs additionally required sufficient trunk control up to the Xyphoid Process. This research was approved by the University of Illinois Institutional Review Board (IRB #23664), with all participants providing informed consent.

Participants' torso maximum range of motion in frontal, sagittal, and transverse planes were measured using the wearable IMU while seated on the stationary PURE. Maximum torso torques in frontal and sagittal planes were recorded using the FSS. mWCUs' that participated in this study came from diverse backgrounds and conditions [20].

TABLE I
Subject Demographics and Anthropometric

| | | ABU | mWCU |
|---|---|---|---|
| Number [Male: Female] | | 10 [5:5] | 10 [4:6] |
| Age (years) | | 23.6 (1.1) | 27.4 (1.6) |
| Height (m) | | 1.68 (0.02) | 1.62 (0.03) |
| Weight (kg) | | 58.1 (2.3) | 51.6 (2.6) |
| Years of using mWC (yrs) | | NA | 19.4 (2.55) |
| Torso Range of Motion (°) | frontal | 63.1 (4.2) | 46.6 (4.5) |
| | sagittal | 58.6 (3.3) | 41.1 (3.5) |
| | transverse | 91.3 (4.2) | 64.5 (9.5) |
| Maximum Torso/FSS Moment Range (Nm) | frontal | 107.1 (8.1) | 82.1 (12.3) |
| | sagittal | 113.1 (7.7) | 85.6 (11.3) |

Values represent mean and (standard error).

### B. Test Course

The HRI study for evaluating the shared-control paradigm included one training course (Fig.3a,d) and two test courses (S-Turn (ST), Fig.3b,e; and Zigzag (ZZ), Fig.3c,f). The training course introduced participants to PURE and its shared-control operation, ensuring familiarity before the testing phase. Distinct designs between the training and test courses prevented participants from memorizing the layout, minimizing learning effects.

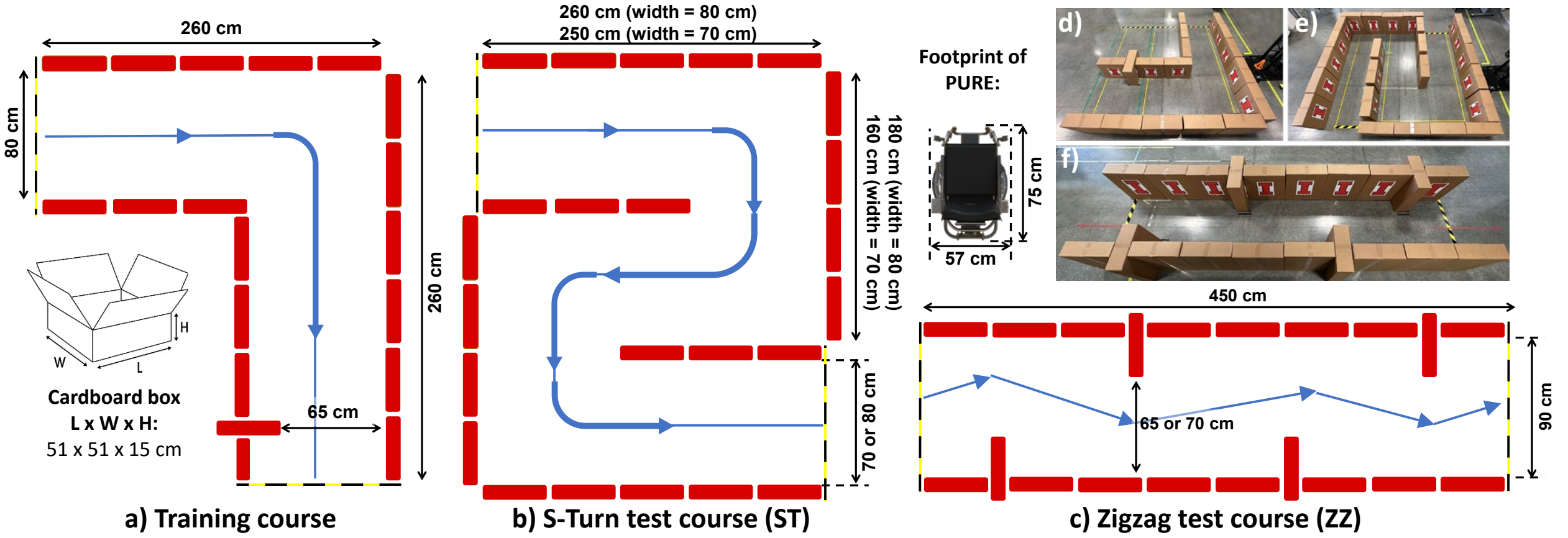

Fig.3. a)-c) The bird's eye view schematics for training course and test courses setup. d)-f) Pictures of the training course and test courses.

The ST and ZZ test courses, simulating challenging indoor scenarios like narrow hallways and static obstacles, had variable widths to accommodate different navigation skill levels. The ST course widths were 70 cm or 80 cm, while the ZZ course had fixed 90 cm paths with 65 cm or 70 cm narrowing sections. Considering PURE's width (~60 cm) and standard wheelchair passage widths (81.5 cm minimum point width, 91.5 cm continuous width as per Uniform Federal Accessibility Standards [22]), these courses presented significant challenges. Width settings were determined through a pilot test with three ABUs and two mWCUs (excluded from the evaluation) to balance difficulty and participant frustration. These settings, more demanding than those in similar studies [2], [8], [9], [10], [11], [23], represented some of the most challenging configurations in the field.

Both training and testing courses used cardboard box walls for a lightweight, discrete structure facilitating collision assessment. Collisions were categorized as "touch" (light contact without dislocating the box), "move" (box pushed out of place), or "failure" (significant disruption preventing course continuation). Only the first instance of collision with each obstacle was recorded. This protocol ensured consistency in collision calculation, and participants were instructed to ignore collided boxes for the remainder of the trial.

### C. Training and Testing

Participants initially familiarized themselves with PURE in an open space and on the training course. They adjusted the TES control sensitivities to suit their movement preferences. After gaining comfort with PURE, the shared-control mode was introduced, accompanied by a briefing on its operational principles. Participants then repeated the training with shared-control activated, customizing parameters like resistive force and alarm volume to their preference. The training concluded once participants felt confident maneuvering PURE in all directions and controllers. Training took ~30 min.

The testing involved navigating PURE from a start to a finish line (marked on the floor, Fig.3d-f) across two test courses (ST and ZZ), each with two different widths. Each participant completed 24 trials in total (2 test courses × 2 control schemes × 2 widths × 3 trials per condition). To counteract potential learning and fatigue biases, the order of test courses and control schemes was randomized. Participants always started with the wider course before moving to the narrower one to reduce initial anxiety. They were instructed to navigate at their comfortable speeds while avoiding collisions. Metrics such as the number of collisions and completion time were captured during each trial.

### D. Metrics and Data Analysis

Inspired by a previous study [24], a Collision Index ($C_i$) was defined by assigning weights to different types of collisions, namely "touch," "move," and "failure." These weights reflect the severity of the collision and the potential risk of structural damage. The $C_i$ is calculated using the following equation:

$$C_i = 1 \times \#\, of\ Touches\ + 3 \times \#\, of\ Moves \\ +9 \times \#\, of\ Failures \qquad (6)$$

The study incorporated three sets of metrics to evaluate various aspects of the system's performance:

1) Objective performance: These included the averaged Collision Index ($C_i$), averaged completion time ($T_c$), averaged number of touch collisions ($C_t$), averaged number of move collisions ($C_m$), and averaged number of failures from three trials per test condition. These metrics provided quantitative measures of performance during the tests.
2) NASA Task Load Index (TLX): The TLX survey [25] was utilized to assess the perceived workload and relative operation effort with and without shared-control. After completing each test course, participants were given a TLX survey. The final TLX scores for each subject were computed by taking the average of the two TLX surveys, similar to the practice in [23].
3) Post-study questionnaire: A post-study questionnaire was designed to collect feedback and preferences regarding various aspects of the shared-control design. The questionnaire utilized a 4-point Likert scale (i.e., "Very Little", "Little", "Much", "Very Much") to gather responses and insights. Participants were also asked to indicate their preferred shared-control paradigm for each testing course. Participants were asked to complete this post-study questionnaire after all their trials.

To examine the impact of using shared-control, two repeated-measures multivariate analysis of variance (RM-MANOVA) tests and one one-way MANOVA test were conducted using IBM SPSS (SPSS Statistics 28.0.1). These tests were performed to answer the research questions regarding the potential effects of shared-control on performance, operational effort, and user feedback, respectively. In the first test, the performance was assessed using the objective metrics: average values for $C_i$, $T_c$, $C_t$, and $C_m$. The average number of failures was not included in the RM-MANOVA because there were only two occurrences during the study (for two ABUs, both in the ST Narrow test course without using shared-control), and these were considered outliers. The operational effort was evaluated using the TLX scores, while the user feedback data were collected using the custom questionnaire. The first RM-MANOVA test included three independent variables: Shared-Control mode (No Shared-Control vs. Shared-Control), test courses (S-Turn Wide (STW), S-Turn Narrow (STN), Zigzag Wide (ZZW), and Zigzag Narrow (ZZN)), and user group (ABU, mWCU). The second RM-MANOVA test evaluated independent variables of shared-control mode and user group. The one-way MANOVA test explored the potential preference differences between subject groups in post-study questionnaire results. Follow-up univariate ANOVA tests were conducted if RM-MANOVA tests reached statistical significance. A significance level of $\alpha = 0.05$ was used for all analyses, except for interaction effects which were modified for interactions of shared-control mode and test courses ($\alpha = 0.05 / 8 = 0.006$).

## IV. Results & Discussion

### A. Objective Metrics

The RM-MANOVA tests on objective metrics revealed significant performance variations based on the use of shared-control and test course settings (Fig.4 and Table II). The performance metrics, excluding average failures due to rarity, were significantly influenced by test course configurations ($p < 0.001$), shared-control utilization ($p < 0.001$), and their interaction ($p = 0.005$), with no notable differences between user groups ($p > 0.05$). Further analysis via univariate ANOVA showed the test course settings significantly affected the average $T_c$, $C_t$, $C_m$, and $C_i$ (all $p < 0.001$). The STN course demonstrated higher difficulty, evidenced by longer $T_c$ (34.3 s) and higher average $C_t$ (1.1), compared to the faster completion (19.6 s) of the ZZW course (Fig.4a). These results highlighted the impact of course width on performance metrics. Shared-control usage notably reduced $C_t$ ($p = 0.002$), $C_m$ ($p < 0.001$), and $C_i$ ($p < 0.001$), with over 50% reduction in $C_i$ across test courses (Fig.4.b). However, it had minimal effect on $T_c$ ($p = 0.99$), indicating that shared-control did not significantly alter travel speed (Fig.4a). Interaction effects between test course and shared-control mode were observed in $T_c$ ($p = 0.027$), $C_i$ ($p = 0.008$), and $C_m$ ($p = 0.034$), suggesting shared-control's enhanced efficacy in cluttered environments. Overall, the performance RM-MANOVA indicates that the shared-control system effectively reduced collision risks without significant slowdowns. This balance is noteworthy, as according to previous research [26], some shared-control systems tend to considerably reduce speed near obstacles.

An insightful qualitative analysis of the shared-control system is provided in Fig.5, showcasing a trial from subject 14, a mWCU with complex spinal conditions (*Spinal Fusion*, *Scoliosis*, *Kyphosis*). Fig.5 illustrates the mWCU's trajectories in both the STN and ZZN courses, color-coded by forward velocity. This example highlights the system's impact on trajectory and speed. Initially, without shared-control, the mWCU's path in the STN course exhibited jittery and erratic movements, with velocities below 0.1 m/s. Upon activating shared-control the trajectory was smoother, and velocities exceeded 0.2 m/s. Similar observations can be found for the ZZN course. These observations underline the shared-control system's effectiveness in reducing necessary motion corrections by offering deceleration assistance: The rider experienced a reduced need for continuous back-and-forth leaning for speed adjustment, maintaining a stable posture while the system managed deceleration and speed control. This aspect is especially beneficial for individuals with limited torso control, simplifying navigation and enhancing overall mobility and comfort.

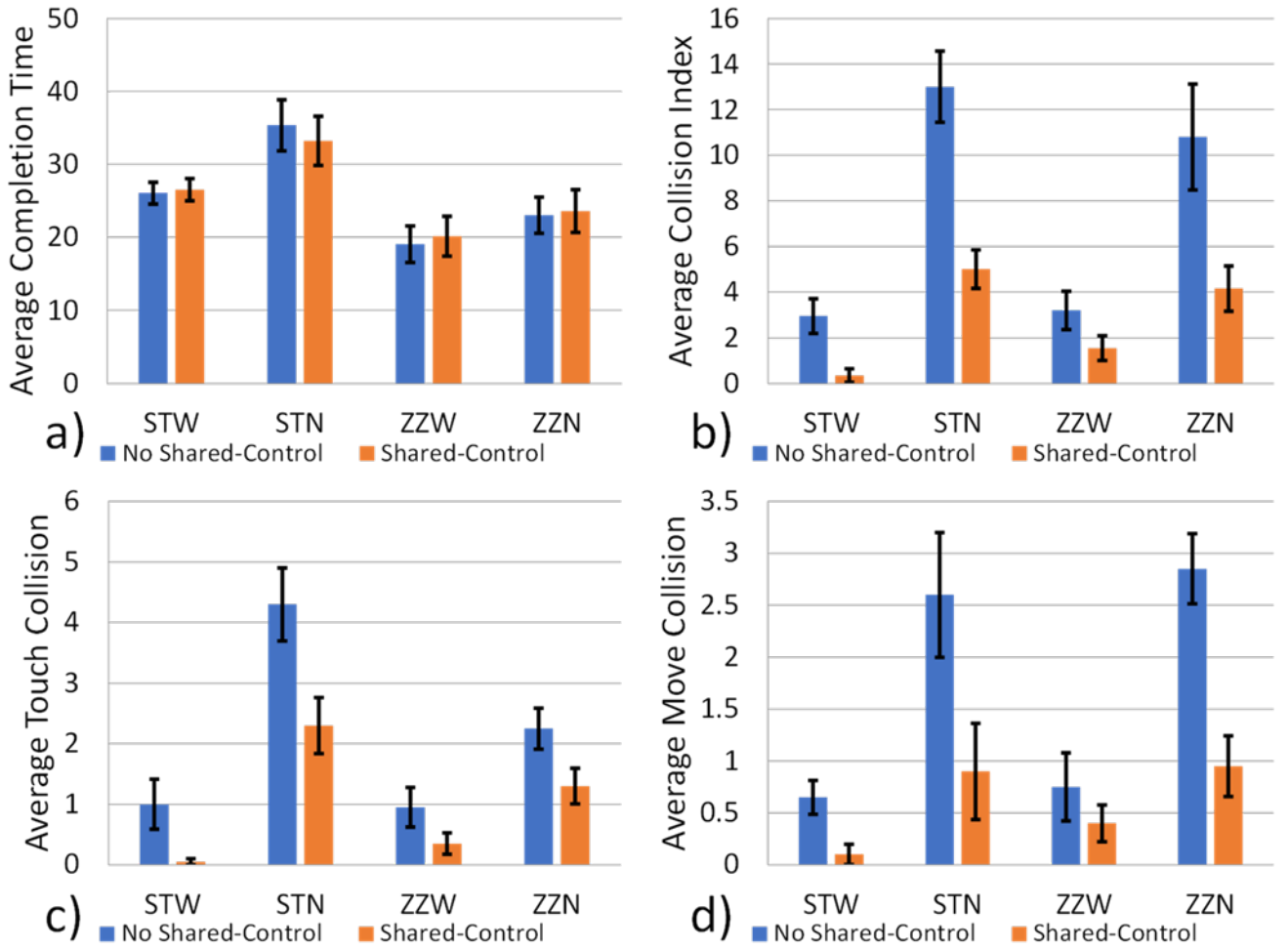

Fig.4. Performance metrics by shared-control mode. a) Average Completion Time. b) Average Collision Index. c) Average Touch Collision. d) Average Move Collision. The error bars represent Standard Errors.

TABLE II
HUMAN-ROBOT INTERACTION TEST OBJECTIVE METRICS

| | Shared-Control Mode | | Test Course | | | |
|---|---|---|---|---|---|---|
| | No S-C (a) | S-C (b) | STW (c) | STN (d) | ZZW (e) | ZZN (f) |
| Avg $T_c$ | 25.9 (2.5) | 25.9 (2.6) | 26.3 (1.5)[de] | 34.3 (3.7)[cef] | 19.6 (2.7)[cdf] | 23.3 (2.8)[de] |
| Avg $C_t$ | 0.7 (0.1)[b] | 0.3 (0.1)[a] | 0.2 (0.1)[df] | 1.1 (0.1)[cef] | 0.2 (0.1)[df] | 0.6 (0.1)[cde] |
| Avg $C_m$ | 0.6 (0.1)[b] | 0.2 (0.1)[a] | 0.1 (0.0)[df] | 0.6 (0.1)[ce] | 0.2 (0.1)[df] | 0.6 (0.2)[ce] |
| Avg $C_i$ | 2.5 (0.4)[b] | 0.9 (0.2)[a] | 0.6 (0.2)[df] | 3 (0.4)[ce] | 0.8 (0.2)[df] | 2.5 (0.5)[ce] |

S-C: Shared Control; Values represent mean and (standard error).
Note: A superscript letter denotes a significant difference from the indicated condition ($p < 0.05$).

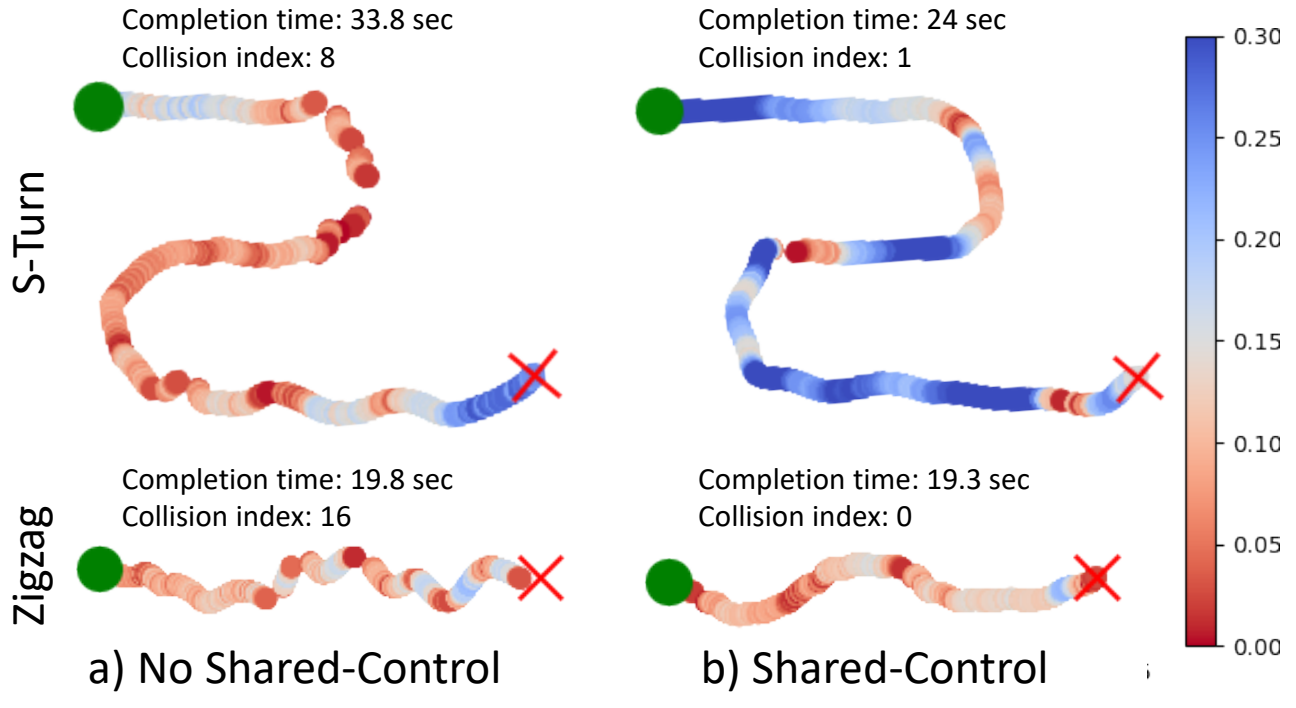

Fig.5. Example trajectories (ST: top, ZZ: bottom) taken by subject 14. Color indicates velocity (m/s) in the forward direction. The "broken" segment indicates PURE drove backward in those instances. The green circles represent starting positions, and the red crosses are where the data collection ended.

### *B. NASA TLX*

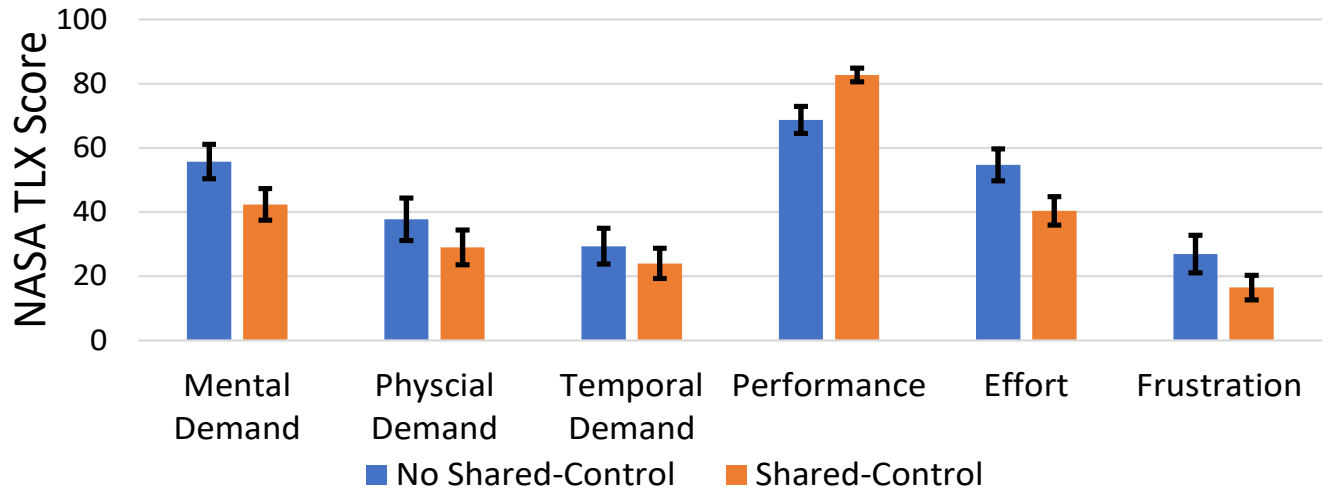

Fig.6. NASA TLX survey results by shared-control mode. The error bars represent Standard Errors.

The operational load RM-MANOVA revealed significant improvements in terms of TLX scores with the use of shared-control. Specifically, there were marked decreases in mental demand ($p = 0.002$), physical demand ($p = 0.022$), effort ($p < 0.001$), and frustration ($p = 0.008$), coupled with an increase in performance scores ($p < 0.001$) (Fig.6). However, the analysis showed no notable differences in these metrics between user groups ($p > 0.05$) and no significant effect on temporal demand ($p = 0.06$). These outcomes indicate that participants experienced a notable reduction in operational effort and an improvement in perceived performance while using the shared-controller. Although no statistical difference was found across user groups, mWCUs tended to report lower scores for mental demand, physical demand, effort, and frustration on average. Overall, the results suggested that the shared-control system contributed to an improvement in their perceived performance. This was further substantiated by the RM-MANOVA findings, which underscored the statistically significant impact of shared-control on reducing cognitive workload and improving user performance.

### *C. Post-study Questionnaire*

The post-study questionnaire aimed to gather insights into participants' preferences regarding the shared-control design. Using a 4-point scale across ten categories (Fig.7a), participants rated the shared-control system. The one-way MANOVA test showed no significant differences in responses between user groups ($p = 0.304$), indicating a general consensus. Participants consistently found the shared-control to be intuitive (average score: 3.55), natural (3.15), and safe (3.5). A noteworthy trend was observed in the responses of manual wheelchair users (mWCUs) and able-bodied users (ABUs). Both groups rated the shared-control as non-aggressive (1.5), with mWCUs tending to rate it as less aggressive than ABUs. ABUs tended to attribute more contribution to haptic feedback in the forward-backward direction compared to mWCUs, but both groups agreed on its importance in the left-right direction. Participants generally found haptic feedback more helpful than audio alarms in test settings and for everyday use. This preference for implicit communication, like haptic feedback, was more pronounced among mWCUs.

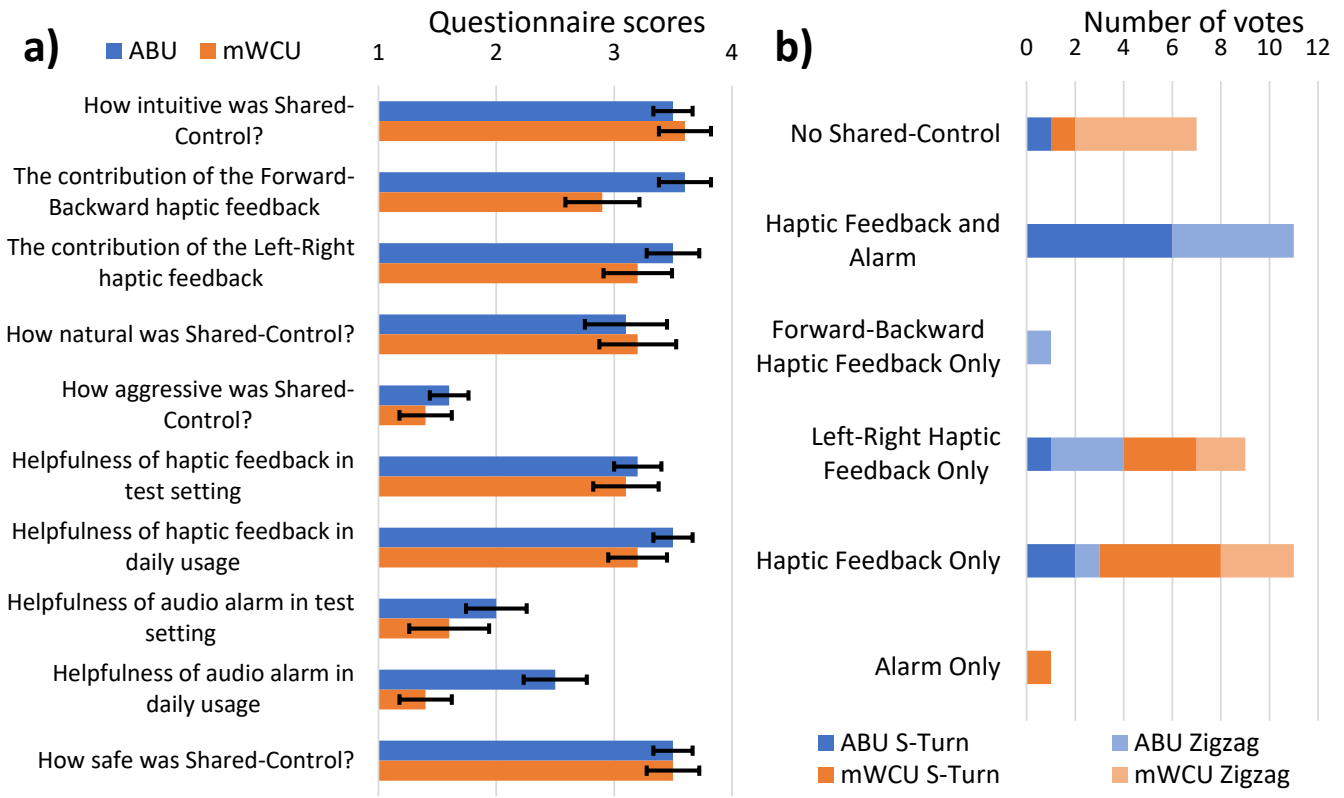

Fig.7. a) Post-Study Questionnaire Scores. The error bars represent Standard Errors. b) Most preferred shared-control paradigm for each test course and subject group.

In a specific query about preferred shared-control paradigms for each test course, over 50% of ABUs favored "Haptic Feedback and Alarm" for both courses, whereas 50% of mWCUs chose "No Shared-Control" for the less challenging ZZ course and "Haptic Feedback Only" for the more demanding ST course. This preference divergence suggests that ABUs valued the added safety of haptic and audio feedback, enhancing their confidence, while mWCUs desired a higher device responsiveness, especially in the ZZ course. This aligns with the broader questionnaire findings (Fig.7b), highlighting a general preference for implicit communication methods, with mWCUs particularly averse to audio alarms, possibly to avoid drawing attention. These results underscore the diversity in shared-control preferences among participants, reflecting the challenge of designing a universal shared-control solution. Similar to findings in [26], it is evident that offering multiple control options is crucial to accommodate individual needs in varied scenarios.

## V. Conclusion & Future Work

This study introduced a novel shared-control approach using Passive Artificial Potential Field (PAPF) to enhance the safety and intuitiveness of a self-balancing mobility device. An experimental study with 20 participants found improved performance and reduced operational effort in tests with manual wheelchair users and able-bodied users. While preferences for shared-control designs varied between groups, no significant differences were found in completion time or collision metrics. This study acknowledges several limitations and areas for future improvement. Future work includes upgrading the single-beam LiDAR to multi-beam LiDAR or RGB-D cameras to enhance obstacle detection, improving the alarm system for better directional cues, and incorporating dynamic, high-speed navigation challenges for thorough shared-control performance assessment. Future research should also explore and compare alternative shared-control paradigms such as the DWA [9] and deliberative shared-control [5], [6], [7] on the PURE platform will provide valuable benchmarks and insights into assistive mobility technology.

## Acknowledgements

The authors thank Yixiao Liu, Keona Banks, Tianyi Han, Zhanpeng Li, Yintao Zhou, and Tommy Nguyen, Maxine He, Jason Robinson, Zhongchun Yu, Prof. Joao Ramos, Prof. Katie Driggs-Campbell, Adam Bleakney, DRES and Center for Autonomy Robotics Lab at UIUC.